\documentclass[letterpaper, 10pt, conference]{ieeeconf}
\IEEEoverridecommandlockouts
\usepackage[utf8]{inputenc}
\usepackage[english]{babel}

\usepackage{comment}
\usepackage{xspace}
\usepackage{graphicx}
\usepackage{overpic}
\usepackage{svg}
\usepackage[dvipsnames]{xcolor}
\usepackage{bm}
\usepackage[hang,flushmargin]{footmisc}
\usepackage{adjustbox}
\usepackage{etoolbox}
\usepackage{environ}
\usepackage{pbalance}
\usepackage{flushend}
\usepackage{mathtools}
\usepackage[normalem]{ulem}

\usepackage{tabularx}
\usepackage{multirow}
\usepackage{booktabs}
\usepackage{colortbl}
\usepackage{float}
\usepackage{placeins}
\usepackage{array}
\usepackage{threeparttable}
\usepackage{makecell}

\usepackage{amsmath}
\usepackage{amssymb}
\usepackage{amsthm}
\usepackage{siunitx}

\DeclareMathOperator*{\argmin}{arg\,min}

\makeatletter
\patchcmd{\@makecaption}{\scshape}{}{}{}
\newcommand{\onetagright}{\tagsleft@false}
\makeatother

\usepackage{subcaption}
\newtheoremstyle{main}
{1em}                                                %
{1em}                                                %
{\itshape}                                           %
{0pt}                                                %
{\scshape}                                           %
{\\*}                                                %
{2pt}                                                %
{\thmname{#1}\thmnumber{ #2}: \thmnote{\itshape #3}} %

\usepackage[linesnumbered,ruled,noend]{algorithm2e}
\newcommand{\removelatexerror}{\let\@latex@error\@gobble}

\let\labelindent\relax
\usepackage[inline]{enumitem}

\usepackage[
	activate   = {true},
	protrusion = false,
	expansion  = true,
	kerning    = true,
	spacing    = true,
	tracking   = false,
	auto       = true,
	selected   = true,
	factor     = 1000,
	stretch    = 10,
	shrink     = 10,
]{microtype}

\usepackage{csquotes}
\usepackage[
	maxbibnames=5,
	maxcitenames=2,
	natbib=true,
	bibstyle=ieee,
	citestyle=numeric-comp,
	backend=biber,
	sorting=none,
	giveninits=true,
	url=false,
	doi=false,
	eprint=false,
	isbn=false,
]{biblatex}

\definecolor{purduegold}{HTML}{C28E0E} %

\makeatletter
\let\NAT@parse\undefined
\makeatother
\usepackage[pdfa,colorlinks,bookmarksopen,bookmarksnumbered,allcolors=purduegold]{hyperref}
\usepackage{bookmark}

\usepackage[nameinlink,capitalise]{cleveref}
\crefname{line}{line}{lines}
\crefname{figure}{Fig.}{Figs.}
\Crefname{figure}{Fig.}{Figs.}
\crefname{equation}{Eq.}{Eqs.}
\Crefname{equation}{Eq.}{Eqs.}
\crefname{section}{Sec.}{Secs.}
\Crefname{section}{Sec.}{Secs.}
\crefname{definition}{Def.}{Defs.}
\Crefname{definition}{Def.}{Defs.}
\crefname{algorithm}{Alg.}{Algs.}
\Crefname{algorithm}{Alg.}{Algs.}
\crefname{assumption}{Asm.}{Asms.}
\Crefname{assumption}{Asm.}{Asms.}
\crefname{subassumption}{Asm.}{Asms.}
\Crefname{subassumption}{Asm.}{Asms.}
\crefname{problem}{Problem}{Problems}
\Crefname{problem}{Problem}{Problems}

\SetKw{InParallel}{in parallel}

\makeatletter
\newcommand\footnoteref[1]{\protected@xdef\@thefnmark{\ref{#1}}\@footnotemark}
\makeatother

\graphicspath{ {./images/} }

\providecommand{\Opt}{\texttt{(Opt)}}

\providecommand{\methodname}{INSPO}

\title{\fontsize{17pt}{24pt}\selectfont \bf Safe Real-Time Policy Steering via Noise-Space Trajectory Optimization for One-Step Generative Policies
}

\newif\ifanonymous
\anonymousfalse

\ifanonymous
  \author{Anonymous Author(s)}
  \hypersetup{hidelinks}
\else
    \author{
    Qingyi Chen, Joseph Ruan, and Zachary Kingston
    \thanks{QC, JR, and ZK are with the Department of Computer Science, Purdue University, West Lafayette, IN, USA. {\tt \{chen5221, ruan44, zkingston\}@purdue.edu}. 
    }}
\fi

\begin{document}
\maketitle
\thispagestyle{empty}
\pagestyle{empty}

\begin{abstract}
Generative robot policies can represent diverse, multimodal behaviors, but adapting pretrained policies to deployment-time constraints such as collision avoidance and orientation maintenance remains challenging. Existing inference-time steering methods typically apply gradient guidance through iterative diffusion or flow processes, which can be computationally expensive for real-time control. We propose INSPO, which formulates inference-time steering of one-step generative policies as trajectory optimization in the policy's input noise space. By optimizing the input noise while evaluating constraints on the induced state trajectory, INSPO searches the policy-induced behavior space without directly modifying generated actions. The optimization includes a regularization term that encourages solutions to remain consistent with the policy's input distribution and is solved online using population-based particle optimization. We evaluate INSPO on state- and image-based task-specific policies and generalist vision-language-action policies across Push-T, Can pick-and-place, and LIBERO-Spatial. INSPO improves task success and constraint satisfaction over best-of-N sampling and action projection, while comparing favorably with gradient-guided generation at lower runtime.
\end{abstract}

\section{Introduction}
\label{sec:intro}

Imitation learning has become a powerful paradigm for developing robot policies from demonstrations, enabling applications ranging from visuomotor manipulation to vision-language-action (VLA) policies~\cite{chi2025diffusion, pmlr-v305-black25a, prasad2024consistency, wang2025onestep}. By learning directly from expert data, these policies can acquire complex skills without explicitly modeling task objectives or robot dynamics. However, policies trained from demonstrations are inherently limited by the conditions represented in the training data. When deployed in unstructured environments in the real world, robots often encounter new requirements that were not present during training such as avoiding obstacles while transporting an object. Enabling a pretrained policy to accommodate new deployment-time constraints without retraining is therefore important for practical robotic deployment.

Generative policies offer a promising framework for such adaptation. Diffusion- and flow-based policies~\cite{chi2025diffusion, ho2020denoising, lipman2023flow, pmlr-v305-black25a, prasad2024consistency, wang2025onestep} model a distribution of task-completing actions instead of learning a single deterministic mapping, capturing diverse behaviors and different ways of completing the same task. This property also allows these policies to be steered at inference time to better satisfy deployment-time objectives, for example, by selecting high-scoring candidates according to task metrics~\cite{ qi2026inference}. A prominent approach is gradient guidance \cite{dhariwal2021diffusion, du2026dynaguide, wang2026ppguide}, which injects gradients of a task cost or reward into the iterative generation process to progressively steer samples toward desired behaviors. While effective, this approach faces several limitations in real-time robotic applications. It requires repeated model evaluations throughout denoising or integration, increasing inference latency and limiting the frequency of closed-loop replanning. Its performance can also be sensitive to the guidance strength and generally does not provide an explicit mechanism to ensure constraint satisfaction. Moreover, as the guidance perturbs intermediate generation steps, strong guidance can push samples away from the learned distribution, leading to a trade-off between constraint satisfaction and task completion. 

\begin{figure}[t]
    \centering
\includegraphics[width=\linewidth]{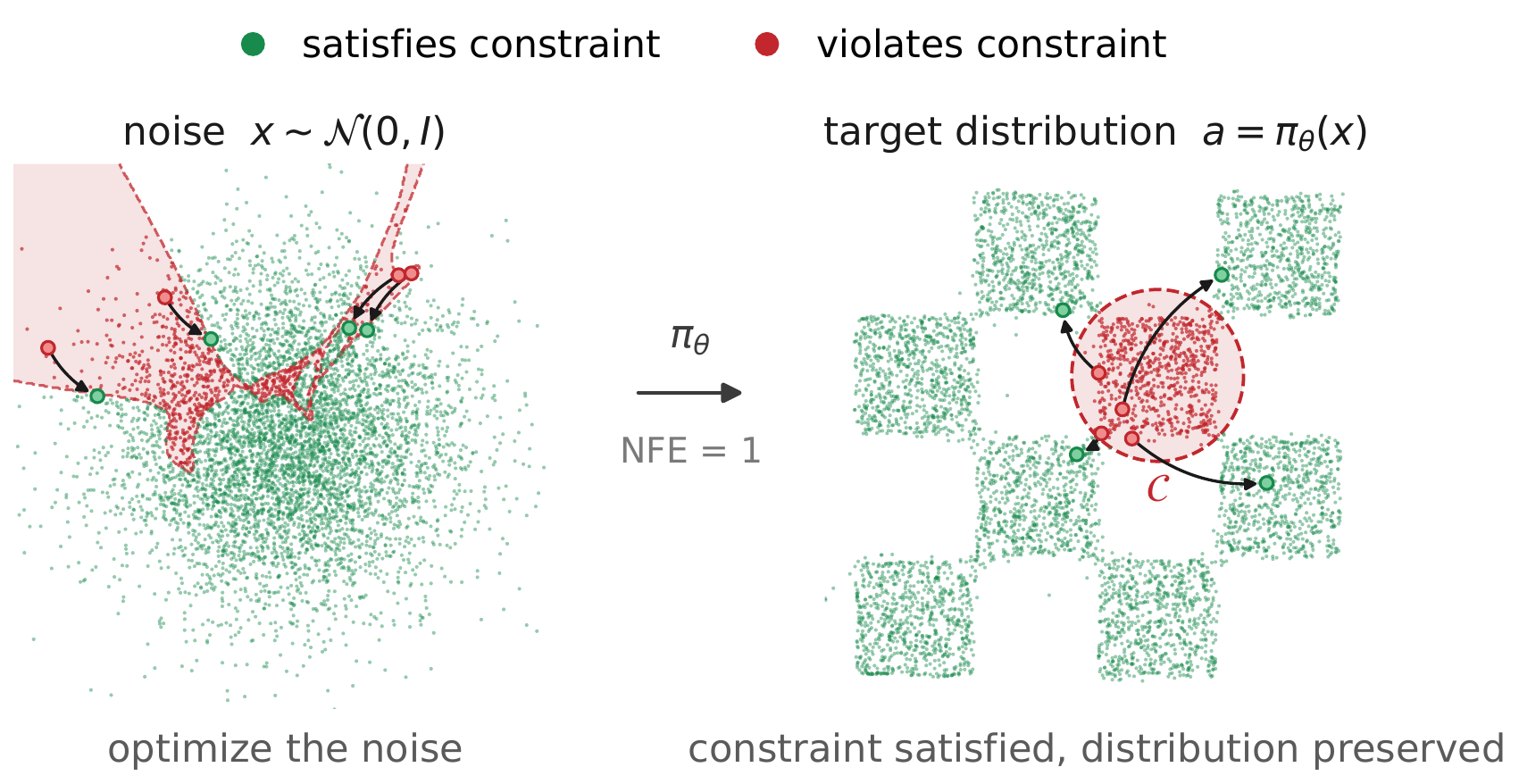}
\caption{\textbf{Inference-time safety steering of a one-step generative policy.} The policy $\pi_\theta$ maps noise samples in the noise space (left) to actions in the action space (right) with a single function evaluation. The red region $\mathcal{C}$ in the action space defines actions that violate the deployment-time constraint. Satisfying and violating actions are respectively shown in green and red, with their corresponding pre-images in the noise space colored accordingly.
At inference time, we optimize the sampled noise to steer violating actions toward satisfying ones while regularizing the noise to remain close to its original distribution.
Arrows show four such steered samples: each links
a violating sample to its optimized counterpart, drawn in both spaces.}
\vspace{-3pt}
\label{fig:intro}
\end{figure}

Recent advances in one-step generative models offer a viable alternative for real-time robotic polices. These models directly map noise samples from a Gaussian distribution to action trajectories, and can generate diverse samples comparable to those of multi-step models using only a single function evaluation~\cite{pmlr-v202-song23a, prasad2024consistency, Yin_2024_CVPR, deng2026generative}. This efficiency is particularly suitable for robotic applications that require frequent replanning and rapid adaptation to changing environments. However, the one-step formulation also removes the intermediate generation process through which conventional gradient guidance performs steering, thus demanding a  new mechanism to incorporate unseen constraints at inference time.

In this work, we propose \methodname{} (\textbf{I}nference-time \textbf{N}oise-\textbf{S}pace \textbf{P}olicy \textbf{O}ptimization), which formulates inference-time steering of one-step generative policies as trajectory optimization in noise space. A pretrained generative policy defines a learned space of task-completing behaviors, where different noise inputs induce different action trajectories. Rather than modifying the generated actions directly, we search over the policy's noise input to select an output behavior that satisfies deployment-time constraints while completing the task. Since a one-step policy maps noise directly to an action trajectory, task objectives and constraints can be evaluated on the resulting trajectory and its rollout while optimization is performed entirely in noise space. This formulation allows the optimizer to leverage behavioral structure learned from demonstrations while adapting to constraints that were not present during training. We demonstrate that our method  enables pretrained one-step policies to adapt to unseen deployment-time constraints and achieves higher success and safety rates over baselines such as action projection, best-of-$N$ selection, and gradient-guidance.

\section{Related Work}
\label{sec:related}

Ensuring that learned robot policies satisfy safety and physical constraints at deployment time has motivated a variety of research. One major class of methods applies safety filters or action projection after generating an action or trajectory, directly modifying the policy output to satisfy constraints~\cite{yao2025inference,pacs2026,yang2026safeflowmatcher}. Other approaches incorporate safety constraints into the generative process. For example, SafeDiffuser~\cite{xiao2025safediffuser} incorporates control barrier functions into diffusion-based planning, while CoBL-Diffusion~\cite{mizuta2024cobl} uses control barrier and Lyapunov functions to guide the denoising process toward safe trajectories. Despite their different mechanisms, these approaches generally enforce constraints either by modifying the generated trajectory or by intervening throughout an iterative generative process.

Beyond explicit safety enforcement, a broader line of work studies inference-time steering of generative policies toward task-specific objectives. Since generative policies model a distribution of behaviors, they can generate multiple candidate trajectories for the same observation, enabling adaptation at deployment time without retraining. A straightforward approach is to sample multiple trajectories and select the candidate with the best task-specific score, commonly referred to as best-of-$N$ selection~\cite{pmlr-v267-beirami25a,qi2026inference}. More active approaches steer samples by modifying the generation process using task-dependent gradients. Originally developed for conditional diffusion generation~\cite{dhariwal2021diffusion}, gradient-based guidance has recently been applied to robot policies to incorporate rewards, dynamics, and other task objectives during generation~\cite{du2026dynaguide,wang2026ppguide}.

A separate line of research aims to reduce the inference latency of generative models by using fewer iterative refinement steps. Consistency models and distillation methods show that generative models can produce diverse samples with substantially fewer function evaluations~\cite{Yin_2024_CVPR, pmlr-v202-song23a, deng2026generative}, and this idea has been extended to robotic control through one-step diffusion and flow-based policies~\cite{prasad2024consistency,li2026one, wang2025onestep}. These approaches substantially reduce action-generation latency, making generative policies more suitable for high-frequency closed-loop control. At the same time, their single-step formulation removes the intermediate generation states typically exploited by gradient-based steering methods. Consequently, inference-time adaptation of one-step policies requires a different mechanism that does not sacrifice their computational advantage.

Recent work has begun to identify the stochastic input of generative models as a useful space for policy adaptation. DSRL~\cite{wagenmaker2025steering} performs reinforcement learning in the latent-noise space of a pretrained diffusion policy to identify favorable regions of the noise distribution, while Golden Ticket~\cite{patil2026youvegotgoldenticket} searches for a fixed initial noise vector that improves downstream task performance. Although both approaches leave the pretrained generative policy unchanged, they rely on an offline adaptation stage to identify favorable noise inputs before deployment. A separate line of work directly optimizes noise for each generation instance. DNO~\cite{karunratanakul2024optimizing} and D-Flow~\cite{pmlr-v235-ben-hamu24a} optimize the initial noise of pretrained diffusion or flow models for controlled generation by backpropagating through the denoising or integration process. Bodmer et al.~\cite{bodmer2026grounding} further formulate the stochastic variables throughout diffusion inference, including the initial noise, as optimization variables for satisfying physical constraints. Such per-instance optimization avoids an additional training stage, but repeatedly differentiating through an iterative generative process incurs substantial computational cost and has primarily been considered for offline generation. In contrast, we perform noise-space optimization directly online at deployment and embed it within a receding-horizon control loop, using a one-step policy to make repeated optimization computationally practical. 

\section{Preliminaries and Problem Statement}
\label{sec:problem}
\subsection{Iterative Generative Policies and Inference-Time Steering}
Diffusion policies formulate robot trajectory generation as a conditional denoising diffusion process~\cite{chi2025diffusion}.  Given the observation history
$\mathbf{O}_t$, a diffusion policy generates the action chunk
$\mathbf{A}_t$ from an initial Gaussian sample
$\mathbf{A}_t^K \sim \mathcal{N}(0,\mathbb{I})$ through $K$ iterative denoising steps:
\begin{align}\label{eq:diffusion}
\mathbf{A}_t^{k-1} = \alpha_k \left(
\mathbf{A}_t^k - \gamma_k\epsilon_\theta(\mathbf{O}_t,\mathbf{A}_t^k,k)+
\mathcal{N}(0,\sigma_k^2 \mathbb{I}
)\right)
\end{align}
for $k=K,\ldots,1$, where $\alpha_k, \gamma_k, \sigma_k$ are noise scheduler parameters, $\epsilon_\theta$ is the learned noise-prediction
network, and the final denoised sample $\mathbf{A}_t^0$ corresponds
to the generated action chunk $\mathbf{A}_t$. Flow-matching policies similarly generate actions by integrating a learned conditional velocity field from an initial noise sample to the action distribution~\cite{lipman2023flow}. In both cases, generating a single action chunk requires multiple evaluations of the network.

The iterative generation process provides a natural mechanism for steering generative policies toward deployment-time objectives. A common approach is to inject gradients of a task-specific objective into intermediate generation steps. For a diffusion policy, let $\ell(\mathbf{A}_t^k)$ denote a task-dependent cost evaluated at an intermediate action representation $\mathbf{A}_t^k$. A generic gradient-guided update can be written as
\begin{equation}
\mathbf{A}_t^{k-1}
=
\mathcal{D}_\theta(\mathbf{A}_t^k,\mathbf{O}_t,k)
-w \nabla_{\mathbf{A}_t^k}\ell(\mathbf{A}_t^k),
\end{equation}
where $\mathcal{D}_\theta$ denotes the original denoising update and $w$ controls the guidance strength. By repeatedly injecting this gradient throughout the denoising process, the generated action can be steered toward desired behaviors.

\subsection{One-Step Generative Policies}

One-step generative models learn a direct mapping from an initial noise distribution to the target data distribution, removing the iterative denoising or integration process at inference time~\cite{prasad2024consistency,deng2026generative,Yin_2024_CVPR}. Let $o_t$ denote the robot's observation at timestep $t$, including its proprioceptive state $s_t$ and other sensor inputs. Let $\pi_\theta$ be a pretrained one-step generative policy parameterized by network weights $\theta$. Given an observation history $\mathbf{O}_t = \{o_{t-T_o+1}, o_{t-T_o+2}, \cdots, o_t\}$ of length $T_o$, the policy estimates the conditional distribution $p(\mathbf{A}_t \mid \mathbf{O}_t)$, where $\mathbf{A}_t$ is an action chunk of length $T_a$. A sample from this distribution is generated by mapping an initial noise variable $x$ to the action chunk
\begin{equation}\label{eq:one-step-generation}
   \pi_\theta(\mathbf{O}_t,x) = \mathbf{A}_t = \{a_{t},  \cdots, a_{t+T_a-1}\},\ x\sim \mathcal{N}(0, \mathbb{I}). 
\end{equation}

In this work, we use Drifting~\cite{deng2026generative} as the underlying one-step generative model. We however note that our steering formulation is not specific to Drifting, and applies to any differentiable policy that maps a stochastic latent variable to an action chunk through a single forward evaluation.

\subsection{Problem Formulation}
At deployment time, the robot is often required to satisfy constraints that were not present during training, such as obstacle avoidance and end-effector orientation maintenance. Executing a generated action chunk $\mathbf{A}_t$ induces a state sequence through the robot dynamics,
\begin{equation}
\label{eq:sys-dynamics}
s_{k+1}=f(s_k,a_k),\quad k=t,\ldots,t+T_a-1.
\end{equation}
over which we define a set of deployment-time constraints $\mathcal{C}=\{c_1,\ldots,c_{N_c}\}$. As the state sequence is determined from the current state $s_t$ and the action chunk $\mathbf{A}_t$ through Equation~\eqref{eq:sys-dynamics}, we write each constraint compactly as $c_j(s_t,\mathbf{A}_t)$. Our goal is to find an initial noise sample $x^\star$ whose generated action chunk completes the task while satisfying all deployment-time constraints such that
\begin{equation}
\label{eq:req}
\tag{Req}
\begin{aligned}
x^\star &\sim \mathcal{N}(0,\mathbb{I}), \\
\mathbf{A}_t^\star
    &= \pi_\theta(\mathbf{O}_t,x^\star), \\
c_j(s_t,\mathbf{A}_t^\star)
    &\leq 0,
    \quad j=1,\ldots,N_c .
\end{aligned}
\end{equation}
As Equation~\eqref{eq:req} requires $x^\star$ to be consistent with the policy's initial Gaussian noise distribution, the resulting action chunk $\mathbf{A}^\star_t$ is generated through the pretrained policy and therefore stays within its policy-induced action space. The remaining constraints further require the induced trajectory to satisfy all deployment-time constraints.

\section{Method}
\label{sec:method}
\subsection{Steering One-Step Policies as Trajectory Optimization}

Although one-step policies expose no intermediate generation states for conventional gradient guidance, their input noise offers a differentiable parameterization of the output. Given a fixed observation history $\mathbf{O}_t$, each initial noise $x$ determines an action chunk through the frozen policy, which in turn induces a state rollout via the robot dynamics in Equation~\eqref{eq:sys-dynamics}. We therefore formulate inference-time steering as trajectory optimization over the policy's input noise space. 

Specifically, we treat $x$ as the optimization variable and transform Requirement~\eqref{eq:req} into an unconstrained objective $\mathcal{J}$ consisting of task and constraint penalties alongside a noise regularization term. Gradients of trajectory-level objectives can be propagated through the system rollout and the one-step policy to update $x$:
\begin{equation}
\nabla_x \mathcal{J}
=
\frac{\partial \mathcal{J}}{\partial \mathbf{A}_t}
\frac{\partial \pi_\theta(\mathbf{O}_t,x)}{\partial x}.
\end{equation}
Following the Gaussian Annulus Theorem~\cite{blum2020foundations},
samples from a high-dimensional standard Gaussian generally concentrate in a shell with radius approximately $\sqrt{\dim(x)}$. We therefore include a noise regularization term that encourages $x$ to stay consistent with the policy's original input distribution using
\begin{equation}
\mathcal{L}_{\mathrm{reg}}(x)
=
\left(
\frac{\|x\|^2_2}{\dim(x)} - 1
\right)^2.
\end{equation}
The resulting optimization problem is
\begin{align}
x^\star = & \argmin_x  \quad
 \mathcal{J}(s_t,\mathbf{A}_t,x)\label{eq:noise-opt}\\
\text{s.t.}\quad
& \mathbf{A}_t=\pi_\theta(\mathbf{O}_t,x),\nonumber\\
& s_{k+1}=f(s_k,a_k),\quad k=t,\ldots,t+T_a-1.\nonumber
\end{align}
where
\begin{equation}
\label{eq:J-def}
\mathcal{J}(s_t,\mathbf{A}_t,x) = \sum_{j=1}^{N_c} \max\left(\lambda_jc_j(s_t,\mathbf{A}_t),0\right) +  \lambda_{\mathrm{reg}} \mathcal{L}_{\mathrm{reg}}(x).
\end{equation}
The first term penalizes violations of the deployment-time
constraints, while $\mathcal{L}_{\mathrm{reg}}$ encourages the
optimized noise to remain consistent with the initial Gaussian distribution.

This formulation differs from directly optimizing the action chunk, as the pretrained policy serves as a parameterization of the candidate behavior space: changing $x$ changes the generated action through the frozen policy rather than directly perturbing the actions. We illustrate this idea in Figure~\ref{fig:intro} using a toy example in which the policy maps a Gaussian noise distribution to a chessboard-pattern target distribution representing task-completing actions. A deployment-time constraint leads part of this target distribution to be infeasible, and our optimization moves the corresponding noise samples toward inputs whose outputs satisfy the constraint. Importantly, because the optimized actions are still generated through the pretrained policy, they remain within the policy-induced target distribution rather than being directly pushed into arbitrary regions of the action space. The noise regularization further discourages solutions far from its nominal input distribution. As $\pi_\theta$ is a one-step policy, each objective and gradient evaluation requires
differentiating through only a single policy evaluation rather than
an iterative denoising or integration process, making repeated
trajectory optimization possible for online receding-horizon
control.

\subsection{Solving the Trajectory Optimization}

We solve the noise-space trajectory optimization using a particle optimization formulation. Following the receding-horizon execution scheme used by Diffusion Policy~\cite{chi2025diffusion}, we distinguish between the action horizon and the execution horizon. At each control step, the policy generates an action chunk of length $T_a$, while only the first $T_e<T_a$ actions are executed before replanning. Similarly, during optimization, we evaluate the constraints over the full action chunk so that the optimizer can account for future behavior, but terminate optimization iterations when the execution chunk satisfies all deployment-time constraints. This allows the optimizer to reason about the future behavior induced by the entire action chunk while reducing planning latency once the actions that will actually be executed are feasible.

At each control step, we optimize a population of $N_p$ noise particles. On the first control step, all particles are sampled from the Gaussian prior; subsequent steps combine warm-started particles from the previous planning cycle with newly sampled particles as described below. Each particle is scored according to the optimization objective $\mathcal{J}$ defined in Equation~\eqref{eq:J-def}. The particles are then updated in parallel via gradient descent at step size $\eta$. We perform at least $K_{\min}$ optimization iterations before allowing early termination even if a feasible execution chunk is found earlier. These additional iterations improve the remainder of the predicted action horizon, providing better solutions for subsequent receding-horizon planning cycles. Optimization terminates after $K_{\min}$ iterations once at least one particle is feasible over the execution horizon, or after at most $K_{\max}$ iterations.

After optimization, we first identify the set of particles whose
execution chunks satisfy all the constraints,
\begin{equation}
\begin{aligned}
\mathcal{I}_{\mathrm{feas}}=\big\{i\in\{1,\ldots,N_p\}
\;\big|\;& c_j
(s_t,\mathbf{A}^{(i)}_{t, 1:T_e})\le 0,  \forall j\big\}.
\end{aligned}
\end{equation}
Among the feasible particles, we select the one that remains
closest to its corresponding original noise sample,
\begin{equation}
\begin{gathered}    
i^\star=\argmin_{i\in\mathcal{I}_{\mathrm{feas}}}
\|x^{(i)}-x_{\mathrm{ref}}^{(i)}\|_2, \\
x^\star = x^{(i^\star)}
\end{gathered}
\vspace{-4pt}
\end{equation}
where $x_{\mathrm{ref}}^{(i)}$ denotes the original unoptimized noise associated with particle $i$. This selection rule
prioritizes constraint satisfaction while favoring the feasible solution that requires the smallest change from the original policy sample.

Since consecutive control steps are strongly correlated, we warm-start the particle population between planning cycles. Specifically, $N_p/2$ of the particles with the lowest cost values from the previous optimization are retained as initial seeds for the next cycle, while the remaining particles are initialized from the Gaussian distribution. The retained particles provide a warm start around previously successful solutions, while the new ones preserve exploration and allow the optimizer to recover from changes in the environment or observation.

The complete procedure is summarized in Algorithm~\ref{alg:noise_optimization}.

\begin{algorithm}[t]
\caption{Online Noise-Space Optimization}
\label{alg:noise_optimization}

\KwIn{
Policy $\pi_\theta$,
population size $N_p$,
minimum iterations $K_{\min}$,
maximum iterations $K_{\max}$,
step size $\eta$,
execution horizon $T_e$
}

Initialize warm-start set
$\mathcal{X}_{\mathrm{warm}} \gets \emptyset$\;

\For{each control step $t$}{
    Get observation history $\mathbf{O}_t$
    and robot state $\mathbf{s}_t$\;

    \eIf{$\mathcal{X}_{\mathrm{warm}} = \emptyset$}{
        Sample reference noise
        $\{x_{\mathrm{ref}}^{(i)}\}_{i=1}^{N_p}
        \sim \mathcal{N}(0,I)$\;
        
        Initialize
        $x^{(i)} \gets x_{\mathrm{ref}}^{(i)}$,
        $\forall i=1,\ldots,N_p$\;
    }{
        Initialize $N_p/2$ particles and their reference noise
        from $\mathcal{X}_{\mathrm{warm}}$\;
        
        Sample reference noise
        $\{x_{\mathrm{ref}}^{(i)}\}$
        for the remaining $N_p/2$ particles from
        $\mathcal{N}(0,I)$\;
        
        Initialize the remaining particles
        $x^{(i)} \gets x_{\mathrm{ref}}^{(i)}$\;
    }

    \For{$k=1,\ldots,K_{\max}$}{
        \For{$i=1,\ldots,N_p$ \InParallel}{
            $\mathbf{A}_t^{(i)}
            \gets
            \pi_\theta(\mathbf{O}_t,x^{(i)})$\;

            Compute
            $\mathcal{J}^{(i)}
            \gets
            \mathcal{J}
            (\mathbf{s}_t,
             \mathbf{A}_t^{(i)},
             x^{(i)})$\;

             $x^{(i)}
                \gets
                x^{(i)}
                -\eta \nabla_{x^{(i)}}\mathcal{J}^{(i)}$\;

            Evaluate constraints over the first
            $T_e$ actions of $\mathbf{A}_t^{(i)}$\;
        }
        \If{$k \geq K_{\min}$ and $\exists$ feasible particle}{
            \textbf{break}\;
        }
    }

    $\mathcal{I}_{\mathrm{feas}}
    \gets
    \{i  \mid
    \mathbf{A}_{t,1:T_e}^{(i)}
    \text{ satisfies all constraints}\}$\;

    $i^\star
    \gets
    \argmin_{{i}\in\mathcal{I}_{\mathrm{feas}}}
    \|x^{(i)}-x_{\mathrm{ref}}^{(i)}\|_2$

    $x^\star \gets x^{i^\star}$;
    
    Execute the first $T_e$ actions of
    $A^\star_{t} = \pi_\theta(\mathbf{O}_t,x^\star)$\;

    Rank particles in ascending order of
    $\mathcal{J}^{(i)}$\;

    $\mathcal{X}_{\mathrm{warm}}
    \gets$
    top $N_p/2$ particle-reference pairs\;
}

\end{algorithm}

\section{Experiments}
\label{sec:experiments}
This section presents four sets of experiments to evaluate \methodname{} on various policy, task, and constraint types. We first study a 2D \textbf{Push-T} task~\cite{florence2021implicit, chi2025diffusion} with obstacle avoidance, where the contact-rich dynamics require the steered action to satisfy the new constraint while preserving the manipulation behavior learned from demonstrations. Second, we consider the manipulation task of \textbf{Can} pick-and-place from robomimic~\cite{robomimic2021} under obstacle-avoidance and end-effector orientation constraints, to simulate manipulation scenarios where the robot must maintain a specified pose during transportation. Third, we apply our method to a VLA policy (i.e., $\pi_{0.5}$~\cite{pmlr-v305-black25a}) on the \textbf{LIBERO-Spatial} benchmark~\cite{liu2023libero}, testing whether our method scales from task-specific policies to large, language-conditioned generative policies. Finally, we validate \methodname{} on a \textbf{real-world lifting} task.

For each task, we consider the following baselines:
\begin{itemize}
    \item \textbf{Drifting/Diffusion Policy~\cite{chi2025diffusion,deng2026generative}}: the pretrained policies without steering. We additionally report their performance without deployment-time constraints in \textcolor{gray}{gray}, providing a reference for their nominal task performance.
    \item \textbf{Best-of-$\mathbf{N}$}: a sampling-based steering strategy that generates $N$ candidate action chunks and executes the one with the lowest cost. We use $N{=}32$ in the experiments.
    \item \textbf{Action projection}: a post-generation optimization that directly modifies the generated action chunk. 
    \item \textbf{Guided Policy}~\cite{dhariwal2021diffusion}: diffusion or flow-matching policies steered within the iterative generative process using gradients of the constraints, under different settings.
\end{itemize}

To provide a comprehensive comparison, we vary the generation and steering settings of the evaluated methods. \texttt{NFE} reports the number of function evaluations used by the generative policy, while \texttt{Param.} denotes the method-specific steering parameter: guidance strength $w$ for Guided Policy and particle count $N_p$ for \methodname{}. We note that NFE denotes the number of sequential generative-network evaluations required to produce each candidate action chunk; parallel samples or particles do not increase this count. For Guided Policy, we evaluate multiple NFE settings and report the best-performing guidance strength $w$ together with values one order of magnitude smaller and larger. For \methodname{}, we evaluate $N_p\in\{8,16,32\}$. Methods without an additional steering parameter are indicated by ``--''.

Performance are evaluated based on the following metrics:
\begin{itemize} 
\item \textbf{Success Rate (SR) / Score:} task success rate for Can and LIBERO-Spatial, and target-region coverage score for Push-T. Episodes containing a constraint violation are counted as failures (zero score).
\item \textbf{Constraint violation Rate (CR):} percentage of episodes containing at least one constraint violation.
\item \textbf{Time:} mean time required for each online control step. 
\item \textbf{Violation:} mean number of execution steps violating the constraints per episode. Unlike CR, which treats any violating episode equally, this metric measures the amount of constraint violations within an episode.
\end{itemize}

Experiments are conducted on a machine with an AMD Ryzen Threadripper PRO 5965WX CPU and an NVIDIA RTX 4090 GPU. We implement the Drifting policies based on the publicly available implementations from Drifting Policy and Drift-VLA~\cite{pham2026driftingpolicy,zuo2026driftvla}.
For image-based task-specific policies, deployment-time obstacles are not rendered into the policy observations, since introducing unseen obstacles shifts the visual  distribution and degrades the pretrained policy's behavior. For the generalist VLA policy on LIBERO-Spatial, we additionally evaluate a setting in which the obstacle is rendered into the policy observation to assess the effect of exposing the policy directly to the deployment-time obstacle.

\subsection{Push-T with Obstacle Avoidance}
We first consider 2D Push-T~\cite{florence2021implicit, chi2025diffusion}, where the robot is tasked to push a T-shaped block to match a traget pose. We consider an obstacle avoidance constraint in addition to the original task, such that the robot is not allowed to pass through the regions defined by the obstacles. As the task is contact-rich, it imposes strong requirement for the steered action to preserve the learned behavior distribution in order not to destroy the subtle contact pattern. We evaluate on both state-based and image-based observations, each composed of 50 scenarios with 2 randomly generated obstacles, and use the same scenarios across all methods for comparison.

\begin{figure}[t]
    \centering
\includegraphics[width=0.96\linewidth]{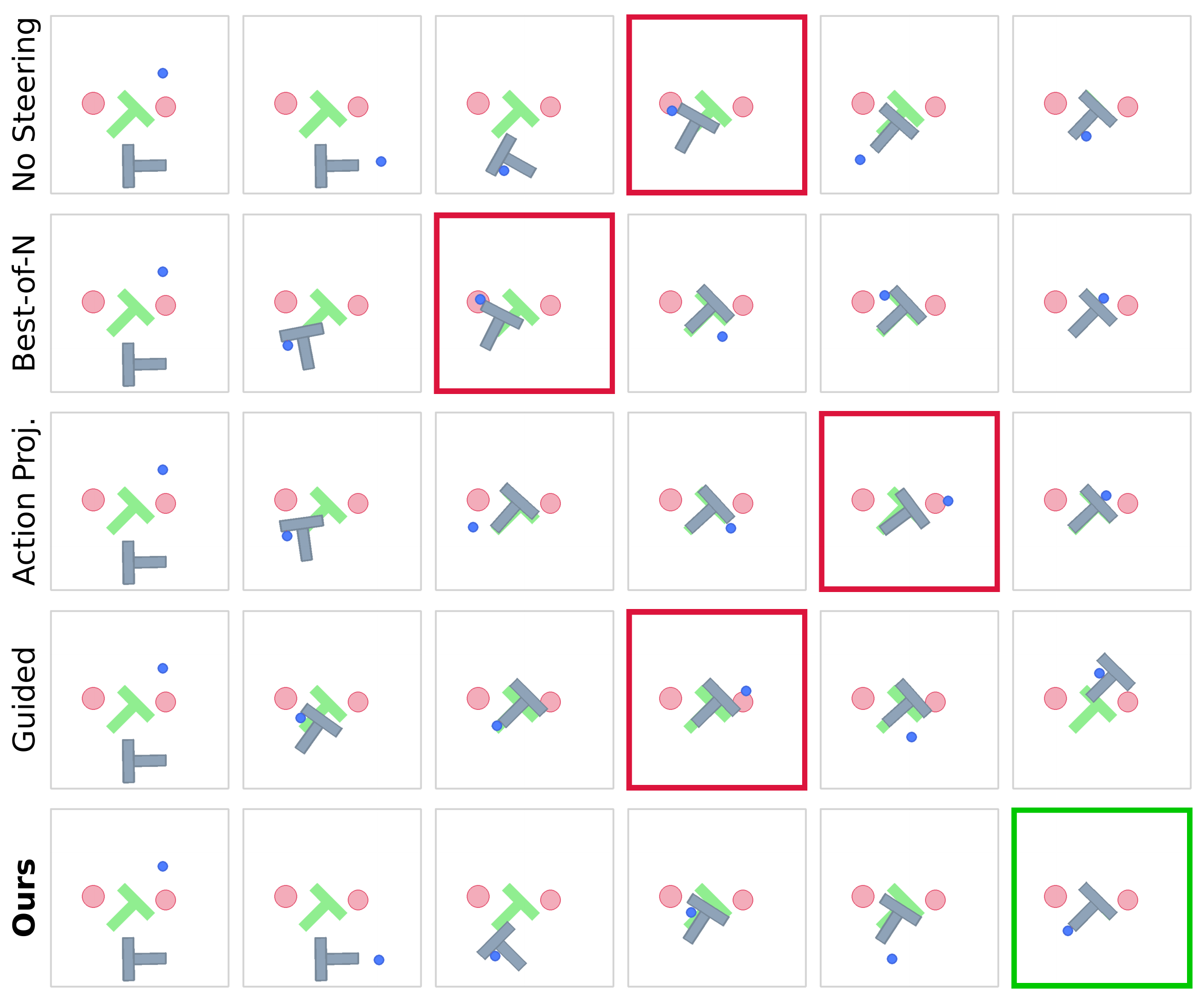}
    \caption{Comparison of trajectories generated by different methods on the Push-T task. Each row shows the progression of one method over time. Red borders indicate collisions, while green borders indicate successful task completion without collision. Corresponding videos are provided in the supplementary materials. }
    \label{fig:pusht}
\end{figure}

Table~\ref{tab:pusht} summarizes the results. First observe that the performance of both unsteered policies significantly drop from their nominal ones, due to the presence of new constraints. Best-of-$N$ provides only limited improvement over the unsteered policies. Action projection substantially reduces constraint violations but still produces collisions in execution. While its projected waypoints are collision-free, direct modification of the action sequence produce motions that are on the boundary of the constraints and can be difficult for the controller to track, leading to constraint violations. The resulting drop in task score further indicates that action-space projection can disrupt the contact-rich behavior required for successful pushing. Gradient guidance achieves a better trade-off between task
performance and constraint satisfaction, but is sensitive to both
the guidance strength and the number of denoising steps. Increasing
the number of function evaluations improves performance, but
at the cost of higher inference time. In contrast, across both observation modalities, \methodname{} achieves the highest task score while completely avoiding constraint violations, with shorter runtime even compared to the ones with NFE=10. Figure~\ref{fig:pusht} shows representative trajectories generated by the different methods. 

\begin{table*}[t]
\centering
\setlength{\tabcolsep}{4pt}
\begin{adjustbox}{max width=\textwidth}
\begin{tabular}{l c l | cccc | cccc}
 & & \multicolumn{1}{c}{} & \multicolumn{4}{c}{State} & \multicolumn{4}{c}{Image} \\
\cmidrule(lr){4-7} \cmidrule(lr){8-11}
Method & NFE & \multicolumn{1}{c}{Param.} & Score $\uparrow$ & CR\% $\downarrow$ & Time [s] $\downarrow$ & \multicolumn{1}{c}{Viol. $\downarrow$}
      & Score $\uparrow$ & CR\% $\downarrow$ & Time [s] $\downarrow$ & Viol. $\downarrow$ \\
\midrule
Diffusion Policy & 100 & \multicolumn{1}{c|}{--} & 18.7 \textcolor{gray}{(92.3)} & 78 & 0.74 $\pm$ 0.02 & 44.28 & 8.4 \textcolor{gray}{(85.6)} & 88 & 0.71 $\pm$ 0.02 & 31.28 \\
\cmidrule(l){2-11}
\multirow{6}{*}{Guided Diffusion} & \multirow{3}{*}{10}
   & $w{=}10^{-6}$ & 29.0 & 66 & 0.09 $\pm$ 0.01 & 16.44 & 31.6 & 58 & 0.09 $\pm$ 0.01 & 4.30 \\
 & & $w{=}10^{-5}$ & 64.3 & 6  & 0.09 $\pm$ 0.01 & 0.14  & 63.6 & 6  & 0.09 $\pm$ 0.01 & 0.18 \\
 & & $w{=}10^{-4}$ & 40.5 & 52 & 0.09 $\pm$ 0.01 & 1.74  & 26.7 & 58 & 0.09 $\pm$ 0.01 & 1.74 \\
\cmidrule(l){2-11}
 & \multirow{3}{*}{100}
   & $w{=}10^{-6}$ & 26.9 & 64 & 0.89 $\pm$ 0.02 & 13.78 & 34.7 & 54 & 0.88 $\pm$ 0.02 & 3.26 \\
 & & $w{=}10^{-5}$ & 65.7 & 2  & 0.89 $\pm$ 0.02 & 0.02  & 64.3 & 12 & 0.90 $\pm$ 0.02 & 0.22 \\
 & & $w{=}10^{-4}$ & 64.0 & 8  & 0.89 $\pm$ 0.02 & 0.12  & 64.3 & 6  & 0.97 $\pm$ 0.07 & 0.06 \\
\midrule
Drifting Policy   & 1 & \multicolumn{1}{c|}{--} & 15.1 \textcolor{gray}{(81.0)} & 76 & \textbf{0.01 $\pm$ 0.00} & 19.90 & 12.6 \textcolor{gray}{(80.9)} & 84 & \textbf{0.01 $\pm$ 0.00} & 26.90 \\
Best-of-$N$ & 1 & \multicolumn{1}{c|}{--} & 28.8 & 54 & \textbf{0.01 $\pm$ 0.01} & 8.72 & 18.0 & 78 & \textbf{0.01 $\pm$ 0.01} & 15.38 \\
Action Projection & 1 & \multicolumn{1}{c|}{--} & 46.6 & 28 & \textbf{0.01 $\pm$ 0.01} & 0.72 & 52.9 & 24 & \textbf{0.01 $\pm$ 0.00} & 0.80 \\
\cmidrule(l){2-11}
\multirow{3}{*}{\methodname{} (ours)} & \multirow{3}{*}{1}
   & $N_p{=}8$  & 62.3 & \textbf{0} & 0.03 $\pm$ 0.04 & \textbf{0.00} & 65.3 & 4 & 0.04 $\pm$ 0.05 & 0.10 \\
 & & $N_p{=}16$ & 66.2 & \textbf{0} & 0.03 $\pm$ 0.04 & \textbf{0.00} & \textbf{70.2} & \textbf{0} & 0.03 $\pm$ 0.04 & \textbf{0.00} \\
 & & $N_p{=}32$ & \textbf{69.3} & \textbf{0} & 0.03 $\pm$ 0.04 & \textbf{0.00} & 67.4 & \textbf{0} & 0.04 $\pm$ 0.04 & \textbf{0.00} \\
\bottomrule
\end{tabular}
\end{adjustbox}
\caption{Obstacle-avoidance results on Push-T with state-based and image-based policies.}
\label{tab:pusht}
\end{table*}

\subsection{Can Pick-and-Place with Manipulation Constraints}
We next evaluate our method on a manipulation task where a Franka Panda manipulator is tasked to pick up a can and place it into a target location using robomimic's Can environment \cite{robomimic2021}. We introduce two deployment-time constraints in addition to the original task objective: the robot must (i) avoid a moving obstacle that is designed to pass through the nominal transport path, and (ii) keep the can upright during transport. This setting therefore evaluates the ability of each method to adapt to multiple constraint types simultaneously. We evaluate each method on 50 trials. In all trials, the obstacle follows the same predefined trajectory, while the initial pose of the can is varied so that the policy must adapt its transport behavior differently across trials. 

\begin{figure}[t]
    \centering
\includegraphics[width=0.96\linewidth]{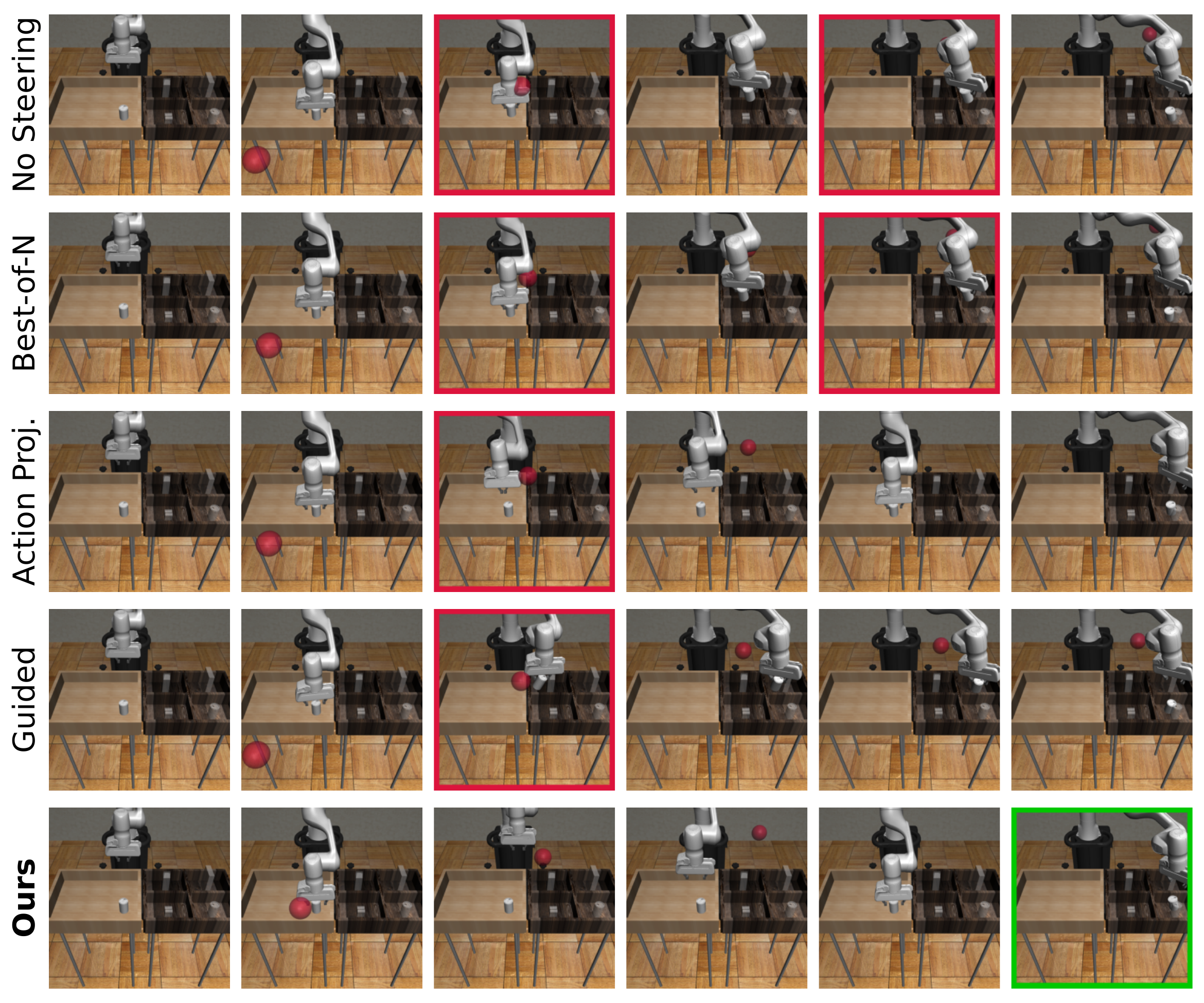}
    \caption{Comparison of trajectories generated by different methods on the Can pick-and-place task. Each row shows the progression of one method over time. Red borders indicate constraint violations, while green borders indicate successful task completion without constraint violations.}
    \label{fig:can}
\end{figure}

Table~\ref{tab:can} reports the results for state-based and image-based policies. Across both observation modalities, \methodname{} provides the most favorable overall balance between task success, constraint satisfaction, and planning efficiency. Guided diffusion becomes competitive when using NFE=100, but the additional denoising steps substantially increase planning time and do not completely eliminate constraint violations. Action projection also performs well on this task and is more effective than in Push-T, suggesting that direct action modification is less disruptive for this manipulation setting; however, occasional collisions remain during execution. In comparison, \methodname{} maintains high task success while reducing constraint violations to near zero and retaining relatively low planning latency. Figure~\ref{fig:can} shows representative trajectories for different methods.

\begin{table*}[t]
\centering
\setlength{\tabcolsep}{4pt}
\begin{adjustbox}{max width=\textwidth}
\begin{tabular}{l c l | cccc | cccc}
 & & \multicolumn{1}{c}{} & \multicolumn{4}{c}{State} & \multicolumn{4}{c}{Image} \\
\cmidrule(lr){4-7} \cmidrule(lr){8-11}
Method & NFE & \multicolumn{1}{c}{Param.} & SR $\uparrow$ & CR\% $\downarrow$ & Time [s] $\downarrow$ & \multicolumn{1}{c}{Viol. $\downarrow$}
      & SR $\uparrow$ & CR\% $\downarrow$ & Time [s] $\downarrow$ & Viol. $\downarrow$ \\
\midrule
Diffusion Policy & 100 & \multicolumn{1}{c|}{--} & 0 \textcolor{gray}{(100)} & 100 & 0.73 $\pm$ 0.02 & 18.76 & 0 \textcolor{gray}{(98)} & 100 & 0.74 $\pm$ 0.02 & 17.88 \\
\cmidrule(l){2-11}
\multirow{6}{*}{Guided Diffusion} & \multirow{3}{*}{10} & $w{=}10^{0}$  & 26 & 74 & 0.12 $\pm$ 0.01 & 8.20  & 18 & 82 & 0.18 $\pm$ 0.03 & 9.76 \\
 & & $w{=}10^{1}$  & 60 & 38 & 0.12 $\pm$ 0.01 & 2.80  & 60 & 32 & 0.17 $\pm$ 0.04 & 1.74 \\
 & & $w{=}10^{2}$  & 34 & 50 & 0.12 $\pm$ 0.01 & 4.82  & 36 & 46 & 0.13 $\pm$ 0.01 & 22.94 \\
\cmidrule(l){2-11}
 & \multirow{3}{*}{100}  & $w{=}10^{0}$  & 30 & 70 & 1.15 $\pm$ 0.02 & 7.96  & 20 & 80 & 1.16 $\pm$ 0.02 & 7.90 \\
 & & $w{=}10^{1}$  & \textbf{90} & 8 & 1.18 $\pm$ 0.02 & 0.54 & 82 & \textbf{2} & 1.19 $\pm$ 0.02 & 0.24 \\
 & & $w{=}10^{2}$  & \textbf{90} & 6 & 1.16 $\pm$ 0.02 & 0.28 & 86 & 10 & 1.16 $\pm$ 0.02 & 0.60 \\
\midrule
Drifting Policy   & 1 & \multicolumn{1}{c|}{--} & 6 \textcolor{gray}{(98)} & 94 & \textbf{0.01 $\pm$ 0.00} & 18.10 & 6 \textcolor{gray}{(98)} & 94 & \textbf{0.01 $\pm$ 0.01} & 18.78 \\
Best-of-$N$ & 1 & \multicolumn{1}{c|}{--} & 12 & 86 & \textbf{0.01 $\pm$ 0.01} & 15.78 & 4 & 96 & 0.02 $\pm$ 0.01 & 20.08 \\
Action Projection & 1 & \multicolumn{1}{c|}{--} & \textbf{90} & 8 & 0.02 $\pm$ 0.04 & 0.36 & \textbf{88} & 12 & 0.05 $\pm$ 0.07 & 0.32 \\
\cmidrule(l){2-11}
\multirow{3}{*}{\methodname{} (ours)} & \multirow{3}{*}{1}
   & $N_p{=}8$  & \textbf{90} & 4 & 0.04 $\pm$ 0.15 & 0.10 & 86 & \textbf{2} & 0.06 $\pm$ 0.14 & 0.08 \\
 & & $N_p{=}16$ & 86 & 2 & 0.04 $\pm$ 0.14 & 0.04 & 86 & \textbf{2} & 0.06 $\pm$ 0.14 & 0.06 \\
 & & $N_p{=}32$ & \textbf{90} & \textbf{0} & 0.04 $\pm$ 0.12 & \textbf{0.00} & \textbf{88} & \textbf{2} & 0.07 $\pm$ 0.19 & \textbf{0.04} \\
\bottomrule
\end{tabular}
\end{adjustbox}
\caption{Results on Can pick-and-place under obstacle-avoidance and orientation constraints, with state-based and image-based policies.}
\label{tab:can}
\end{table*}

\begin{figure}[bt]
    \centering
\includegraphics[width=0.96\linewidth]{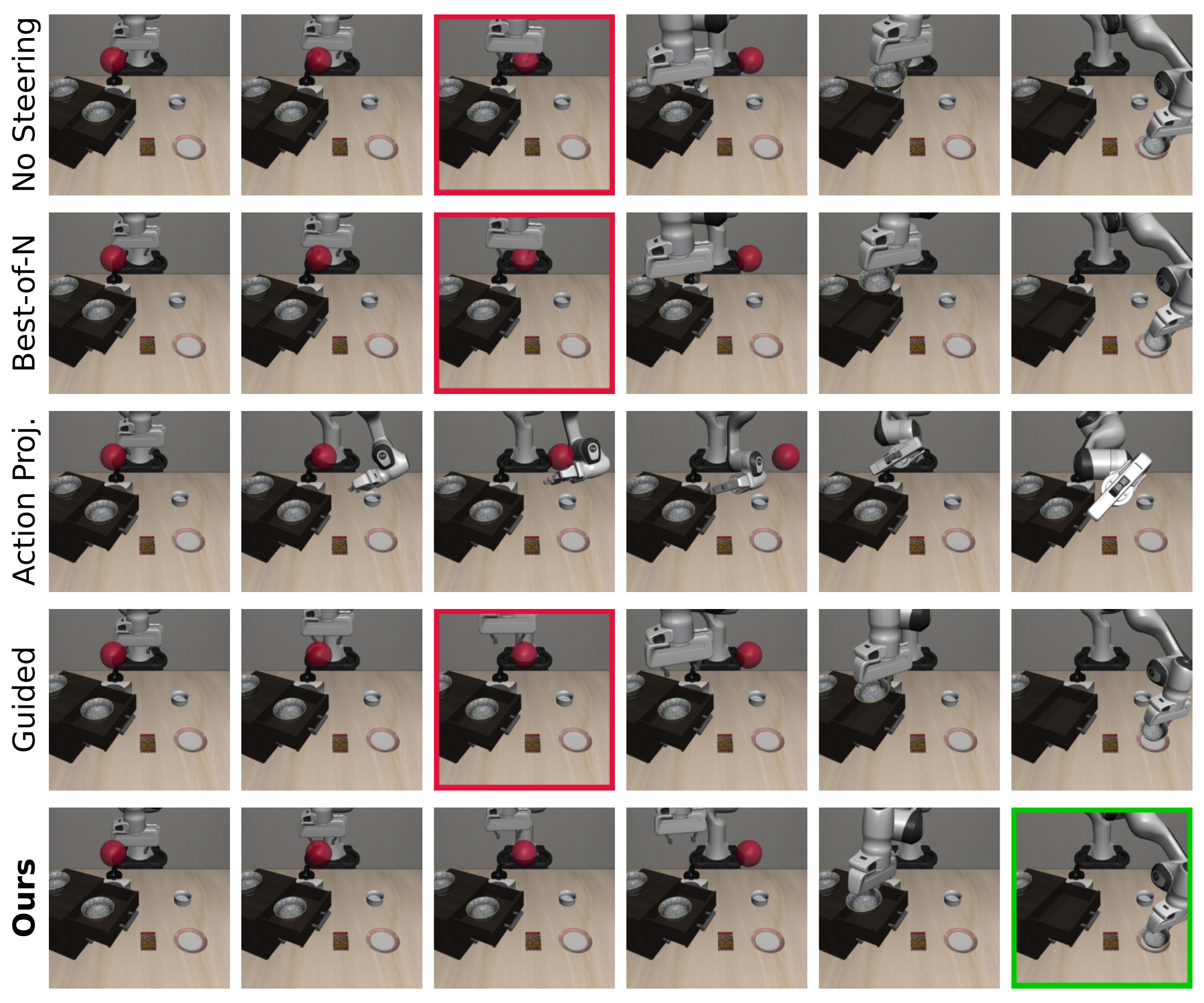}
    \caption{Comparison of trajectories generated by different methods on the LIBERO-Spatial task. Each row shows the progression of one method over time. Red borders indicate collisions, while green borders indicate successful task completion without collision.}
    \label{fig:libero}
\end{figure}

\subsection{VLA Steering for Obstacle Avoidance}
We further evaluate our method on the LIBERO-Spatial benchmark~\cite{liu2023libero} using the
$\pi_{0.5}$ VLA policy~\cite{pmlr-v305-black25a}. The benchmark contains 10 manipulation tasks, each requiring a Franka Panda robot to move an object to a target location specified by a language instruction. This experiment evaluates whether the proposed approach extends beyond task-specific policies to a
large, language-conditioned generalist policy. For each task, we design an obstacle trajectory that intersects
the nominal task-completing motion and therefore requires the policy to adapt its behavior during execution. The task configuration, including object and target locations, is fixed across random seeds. We evaluate each of the 10 tasks using five
different seeds to account for stochasticity in policy execution, for a total of 50 trials. Following the standard inference setting of $\pi_{0.5}$~\cite{pmlr-v305-black25a}, we use NFE=10 for gradient guidance and omit the NFE=100 variants considered in the previous experiments.

Table~\ref{tab:libero} summarizes the results. Across the 10 LIBERO-Spatial tasks, \methodname{} still provides the most favorable overall balance between task success, constraint satisfaction, and planning efficiency. Gradient guidance substantially improves results over the unsteered $\pi_{0.5}$ policy, but
continues to exhibit a noticeable gap in constraint satisfaction across the tested guidance strengths. Action projection also performs competitively in avoiding obstacles, but incurs substantially higher and more variable planning time and retains occasional collisions during execution. Best-of-$N$ provides
little improvement over the unsteered one-step policy. These results suggest that noise-space trajectory optimization can extend to large VLA policies while retaining the favorable inference efficiency of the one-step formulation. Figure~\ref{fig:libero} shows representative executions of different methods. Notably, although Action Projection avoids collision in this example, directly modifying the action drives the robot into an out-of-distribution state, causing the episode to time out.

\subsection{Real-World Validation}
\begin{figure*}
    \centering
    \includegraphics[width=0.95\linewidth]{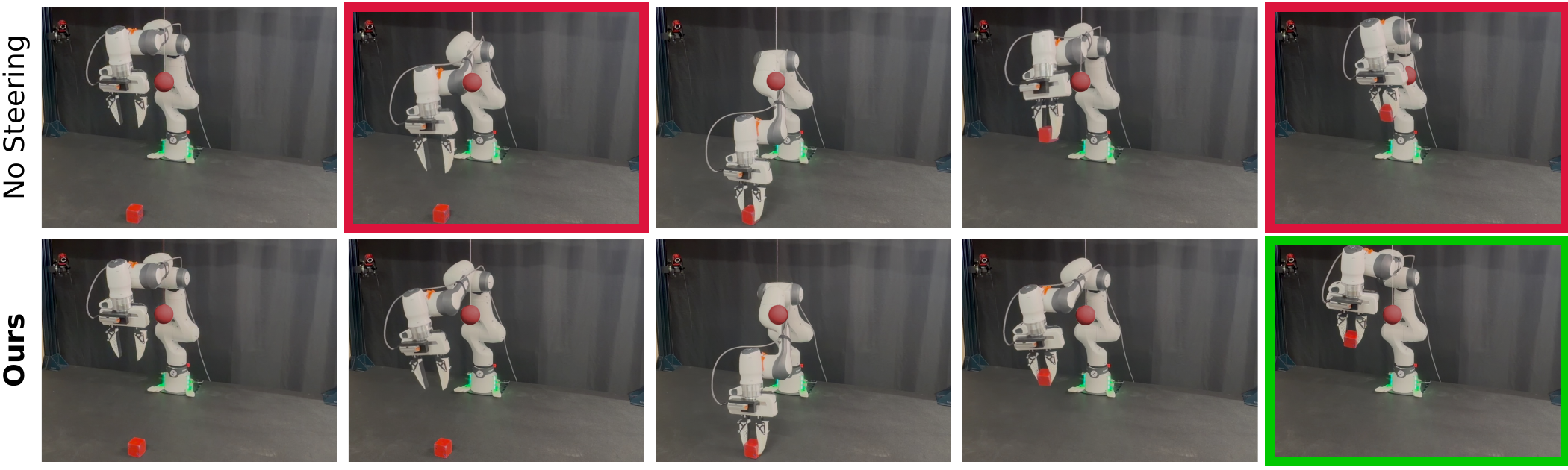}
    \caption{Validation on a real-world lifting task with an obstacle (dark red sphere) introduced only at deployment. Each row from left to right is one method over time. \textbf{Top:} without steering, the robot collides with the obstacle (red borders). \textbf{Bottom:} our method steers the robot around the obstacle and completes the lift (green border). The obstacle is re-colored from white for clarity of presentation. Corresponding videos are provided in the supplementary materials.}

    \label{fig:real-world}
\end{figure*}

Finally, we validate \methodname{} on a real-world lifting task. A Franka Research 3 robot is tasked to lift a block to a target height while avoiding an obstacle introduced at deployment time. We collect 55 obstacle-free demonstrations to train a Drifting Policy and compare \methodname{} against the unsteered policy and Best-of-\(N\) sampling. We omit Action Projection because its optimization frequently exceeds our planning-time limit, preventing reliable real-world execution. Table~\ref{tab:real-world} reports results over 10 trials per method. \methodname{} succeeds in 8/10 trials with only 2 collisions, compared with 4/10 successes for Best-of-\(N\) and 2/10 for the unsteered policy, while retaining low planning latency. Figure~\ref{fig:real-world} shows an example of our method completing the task without collision.

\begin{table}[bt]
    \centering
    \begin{tabular}{l c c c}
    \toprule
     Method & \# Success $\uparrow$ & \# Collision $\downarrow$ & Time [s] $\downarrow$ \\ \midrule
     Drifting Policy & 2 & 8 &  \textbf{0.01 $\pm$ 0.01} \\
     Best-of-$N$ & 4 & 6 & 0.02 $\pm$ 0.01\\
     \methodname{} (ours) & \textbf{8} & \textbf{2} & 0.06 $\pm$ 0.10\\
     \bottomrule
    \end{tabular}
    \caption{Obstacle Avoidance results in real-world lifting experiments.}
    \label{tab:real-world}
\end{table}

\begin{table*}[t]
\centering
\setlength{\tabcolsep}{4pt}
\begin{adjustbox}{max width=\textwidth}
\begin{tabular}{l c l | cccc | cccc}
 & & \multicolumn{1}{c}{} & \multicolumn{4}{c}{Not Rendered} & \multicolumn{4}{c}{Rendered} \\
\cmidrule(lr){4-7} \cmidrule(lr){8-11}
Method & NFE & \multicolumn{1}{c}{Param.} & Score $\uparrow$ & CR\% $\downarrow$ & Time [s] $\downarrow$ & \multicolumn{1}{c}{Viol. $\downarrow$}
      & Score $\uparrow$ & CR\% $\downarrow$ & Time [s] $\downarrow$ & Viol. $\downarrow$ \\
\midrule
$\pi_{0.5}$ & 10 & \multicolumn{1}{c|}{--} & 4 \textcolor{gray}{(98)} & 96 & 0.40 $\pm$ 0.03 & 16.58 & 14 \textcolor{gray}{(98)} & 86 & 0.45 $\pm$ 0.04 & 15.54 \\
\cmidrule(l){2-7}\cmidrule(lr){8-11}
\multirow{3}{*}{Guided $\pi_{0.5}$} & \multirow{3}{*}{10}
   & $w{=}10^0$   & 54 & 40 & 0.57 $\pm$ 0.04 & 7.12 & 46 & 46 & 0.68 $\pm$ 0.05 & 7.92\\
 & & $w{=}10^1$  & 50 & 14 & 0.56 $\pm$ 0.04 & 6.78 & 62 & 14 & 0.67 $\pm$ 0.05 & 6.80\\
 & & $w{=}10^2$ & 60 & 18 & 0.57 $\pm$ 0.03 & 5.74 & 58 & 16 & 0.67 $\pm$ 0.05 & 6.06 \\
\midrule
Drifting Policy   & 1 & \multicolumn{1}{c|}{--} & 8 \textcolor{gray}{(100)} & 92 & \textbf{0.12 $\pm$ 0.03} & 15.78  & 8 \textcolor{gray}{(98)} & 92 & \textbf{0.12 $\pm$ 0.03} & 15.02 \\
Best-of-$N$ & 1 & \multicolumn{1}{c|}{--} & 6 & 94 & 0.15 $\pm$ 0.04 & 15.78 & 14 & 86 & 0.15 $\pm$ 0.04 & 14.74 \\
Action Projection & 1 & \multicolumn{1}{c|}{--} & 60 & 14 & 2.22 $\pm$ 5.39 & 1.68 &  64 & 16 & 11.93 $\pm$ 30.64 & 2.42 \\
\cmidrule(l){2-7}\cmidrule(l){8-11}
\multirow{3}{*}{\methodname{} (ours)} & \multirow{3}{*}{1}
   & $N_p{=}8$  & \textbf{86} & \textbf{2} & 0.43 $\pm$ 0.64 & \textbf{0.62} & \textbf{90} & \textbf{2} & 0.50 $\pm$ 0.77 & \textbf{0.70} \\
 & & $N_p{=}16$ & 80 & 6 & 0.47 $\pm$ 0.70 & 1.06 & 80 & 4 & 0.49 $\pm$ 0.76 & 1.50\\
 & & $N_p{=}32$ & 82 & \textbf{2} & 0.48 $\pm$ 0.76 & 0.74 & 82 & 6 & 0.51 $\pm$ 0.79 & 0.88 \\
\bottomrule
\end{tabular}%
\end{adjustbox}
\caption{Obstacle-avoidance results on LIBERO-Spatial, with the obstacle rendered and not rendered into the policy's observation.}
\label{tab:libero}
\end{table*}

\section{Conclusion}
This work presents \methodname{}, an inference-time steering framework that formulates adaptation of one-step generative robot policies as trajectory optimization over their initial noise. By optimizing the noise rather than directly modifying the generated actions, our method selects from the pretrained policy's induced action space a trajectory that is both task-completing and constraint-satisfying. The one-step formulation enables this optimization to be performed efficiently within a receding-horizon control loop. Across Push-T, Can pick-and-place, and LIBERO-Spatial, our experiments show that \methodname{} consistently provides a favorable balance between task completion, constraint satisfaction, and planning efficiency compared with existing methods.

Future work aims to address several current limitations. First, as optimization is performed through the frozen pretrained policy, it is constrained by the behaviors represented by the learned policy and cannot exhibit behaviors that are significantly out of the training distribution. Second, satisfying constraints on the generated waypoints does not guarantee constraint satisfaction during execution, as tracking errors and controller dynamics may cause the executed trajectory to deviate from the planned trajectory. When deployment-time constraints have no feasible solution within the behavior space of the pretrained policy, fail-safe mechanisms are required. Finally, our current implementation uses first-order gradient descent to solve the noise-space trajectory optimization. More advanced optimization techniques may help improve convergence and computational efficiency. Future work aims to address these challenges and extend this framework to broader classes of constraints and policies.
\label{sec:conclusion}

\ifanonymous
\else
\fi

\printbibliography{}

\end{document}